\documentclass[runningheads]{llncs}
\usepackage{bm}
\usepackage{amsfonts}
\usepackage[T1]{fontenc}
\usepackage{amsmath, amssymb}
\usepackage{cleveref}
\usepackage{wrapfig}
\usepackage{graphicx}
\usepackage{subfigure}
\DeclareMathOperator*{\argmax}{arg\,max}

\newcommand{\fm}{f_{\texttt{SAM}}}
\begin{document}
\title{Task-driven Prompt Evolution for Foundation Models}
%
%
\author{Rachana Sathish\inst{1} \and
Rahul Venkataramani\inst{1} \and
K S Shriram\inst{1}\and
Prasad Sudhakar\inst{1}}
\authorrunning{R. Sathish et al.}
%
\institute{GE HealthCare\\
\email{\{rahul.venkataramani\}ge.com}}
\maketitle              
\begin{abstract}
Promptable foundation models, particularly Segment Anything Model (SAM)  \cite{kirillov2023segment}, have emerged as a promising alternative to the traditional task-specific supervised learning for image segmentation. However, many evaluation studies have found that their performance on medical imaging modalities to be underwhelming compared to conventional deep learning methods. In the world of large pre-trained language and vision-language models, learning prompt from downstream tasks has achieved considerable success in improving performance. In this work, we propose a plug-and-play \textbf{P}rompt \textbf{O}ptimization \textbf{T}echnique for foundation models like \textbf{SAM} (SAMPOT) that utilizes the downstream segmentation task to optimize the human-provided prompt to obtain improved performance. We demonstrate the utility of SAMPOT on lung segmentation in chest X-ray images and obtain an improvement on a significant number of cases ($\sim75\%$) over human-provided initial prompts. We hope this work will lead to further investigations in the nascent field of automatic visual prompt-tuning. 
\end{abstract}
%
%
\section{Introduction}
\label{sec:intro}
The recent release of a foundation model for image segmentation called Segment Anything (SAM) \cite{kirillov2023segment} has generated unprecedented excitement about the possibility of realizing artificial general intelligence (AGI) in the field of medical image analysis. SAM is a task-agnostic promptable segmentation model trained on 1 billion masks. This has triggered the possibility of improved zero-shot segmentation performance and obviate the necessity for specialized techniques across medical imaging tasks \cite{li2022domain}.

Consequently, a number of studies \cite{he2023accuracy,cheng2023sam,ma2023segment} have evaluated the performance of SAM on a plethora of medical imaging segmentation tasks, and have concluded that while SAM is a promising first step, there exists a significant gap compared to supervised learning algorithms on many datasets. The hypothesized reasons include lack of medical imaging samples in the training database and peculiarities associated with medical images (e.g., scan-cone in Ultrasound, 3D nature of CT/MR, large intensity variations in X-Ray and higher image resolution compared to natural images).
 
This sub-optimal performance has prompted researchers to fine-tune the models to medical imaging modalities using parameter-efficient techniques like Low-rank adaptation (LoRA) \cite{zhang2023customized,ma2023segment} and Adapters \cite{wu2023medical} . However,  given the size of networks, fine-tuning these models also requires access to large scale medical image and label pairs.  Obtaining such large scale datasets and availability of heavy compute is beyond the scope of most small research organizations, thereby limiting the adoption of SAM.

An alternate direction to improve the performance on downstream tasks is to learn efficient prompts tailoring for the tasks. A number of works like CoOp  \cite{zhou2022learning}, CoCoOp \cite{zhou2022conditional} have demonstrated the benefit of learning prompts to adapt CLIP-like vision-language models for downstream tasks. Prompt learning not only improves performance over hand-crafted prompts but also reduces manual effort and expertise required in designing the prompts. While these techniques have been explored extensively in natural language processing and vision-language community, their utilization for optimizing prompts for foundation segmentation models has been conspicuously absent.

In this paper, we present a prompt learning method for segmentation foundation models, and demonstrate it on the task of left-lung segmentation on chest X-ray images. To demonstrate the challenges involved and motivate the need for prompt learning, we compute the sensitivity of SAM's output to the choice of prompt's spatial location.

\begin{wrapfigure}{l}{0.4\textwidth}
\centering
\includegraphics[width=0.38\textwidth]{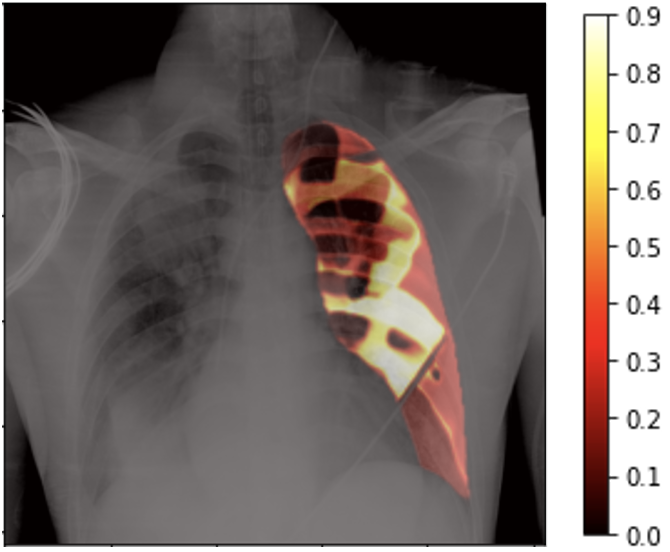}
\caption{Heat-map of Dice values obtained by placing the prompt at various locations in the lung.} \label{heatmap}
\end{wrapfigure}
~\Cref{heatmap} shows the overlay of a chest X-ray image and the heat-map of Dice values when the prompt is placed at different locations of the lung region. The large diversity of Dice values (0.2 to 0.9) highlights that given a click prompt inside the lung region of an X-ray image, it is plausible that another location provides a more accurate segmentation.

Since X-ray is a summative modality, the intensity values under the lung mask are a result of superimposition of soft tissue, ribs, cardiac region, and occasional extraneous objects such as PICC lines. Though visually the lung region may appear equally dark in X-ray images to the user, it is not homogeneous, and its heterogeneity is further amplified by the presence of pathology.

\subsection{Our Approach} 

To improve the segmentation performance in such confounding settings, we propose a \textbf{p}rompt \textbf{o}ptimization \textbf{t}echnique (SAMPOT) that utilizes the knowledge of the downstream task to optimally locate the human-provided prompt to obtain a better segmentation output. We design an unsupervised segmentation performance scorer that generates a proxy for the supervised performance metric like the Dice value. At inference, given a test image and prompt, we iteratively maximize this task-based score to \emph{evolve} the location of the prompt to produce superior results compared to utilizing initial prompt location provided by user. Although we develop this method on SAM, SAMPOT can be used in a plug-and-play fashion with any foundation segmentation model. \\

\subsection{Contributions}
\label{subsec:contrib}
\begin{enumerate}
    \item We propose a plug-and-play prompt optimization technique, SAMPOT, for any promptable segmentation algorithm which fine-tunes an input prompt. To the best of our knowledge, this is the first instance of an automatic prompt tuning strategy for foundation segmentation models.
    \item We demonstrate the efficacy of SAMPOT on the task of segmenting lungs in chest X-ray images and achieve segmentation gains on $\sim75\%$ of the test images. 

\end{enumerate}


\section{Methodology}
\label{sec:method}
We shall introduce a few relevant notations before presenting the method.

\noindent
\textbf{SAM Model:} Let us denote the SAM under consideration by $\fm$, a very large deep neural network model that takes an image $\mathit{X} \in \mathbb{R}^{N \times N}$ and a prompt $\bm{p}$ as input to predict the segmentation mask $\widehat{Y} := \fm(X, \bm{p}) \in \mathbb{R}^{N \times N}$.

\noindent
\textbf{Prompt:} 
For segmentation foundation models such as SAM, a prompt can be a point coordinate, bounding box, dense segmentation, or a text input. It is typically accompanied by a label which indicates whether the prompt is in the foreground $(1)$ or otherwise $(0)$. While SAM can simultaneously take a set of heterogeneous prompts, in this work, we consider one single coordinate prompt $\bm p = (x, y, c)^\intercal, \; x,y\in[N]:=\{0,1,\cdots,N-1\},\;c\in\{0,1\}$. We assume that the prompt is provided by a human user at the start, and it always lies in the foreground object of interest ($c=1$). Therefore, without loss of generality, we can consider $\bm p$ to be a two-component vector representing the 2D coordinates. 

\subsection{Prompt optimization by oracle scoring}
\label{subsec:prompt_optimization}
Our method is aimed at evolving the location of the prompt and arriving at an optimal prompt $\bm p^*$. Suppose we had access to the ground truth mask $Y_{\text{test}}$ for a given input image, we could simply compute the loss $\mathcal{L}_{\text{task}}(\widehat{Y}_{\text{test}} , Y_{\text{test}})$ and choose a $\bm p$ that minimises the loss. However, as that is fallaciously self-fulfilling, we propose to use an oracle $\mathcal O$ that acts as a surrogate to the true loss $\mathcal{L}_{\text{task}}$. The scorer takes the input image $X_{\text{test}}$ and the predicted mask $\widehat Y_{\text{test}}$ and produces a score $s$. The scorer can be a pre-learnt (and fixed) neural network model that can be used in conjunction with the segmentation model, enabling us to compute the gradients of the score with repect to $\bm p$. If the scorer is designed to be positively correlated to the performance metric, we can then solve the following maximization problem to achieve our objective:
\begin{align}
\label{eq:prompt-tuning}
\bm{p}^{\ast} &:= \argmax_{\bm{p}}\mathcal O(X_{\text{test}}, \widehat{Y}_{\text{test}}), \; \text{where}\; \widehat{Y}_{\text{test}} := \fm(X_{\text{test}} , \bm{p}).
\end{align}
Note that the gradient of $s$ is computed with respect to $\bm p$ and therefore only $\bm p$ gets updated, while the weights of SAM $\fm$ and the scorer $\mathcal O$ are held fixed. 

\subsection{Learning to score}
\label{subsec:learning_to_score}
The oracle $\mathcal O$ is expected to score the quality of segmentation blindly in the absence of ground truth. To this end, we train a \emph{segmentation regressor} which learns to predict the Dice directly from the input image and the corresponding predicted mask. This segmentation regressor is trained using a small dataset of input images and ground truth masks. For every input image, several candidate masks are synthetically generated by modifying the true segmentation mask, and their corresponding Dice coefficients are computed. This extended set of images, masks and Dice scores are then used to train the regressor. The details of candidate mask generation and segmentation regressor are described in~\cref{sec:seg_reg}. In general, segmentation quality score can be vector valued and along with the described regressor, one can use adversarial loss~\cite{luc2016semantic}, shape autoencoder~\cite{ravishankar2017learning}, etc. 

\begin{figure}[t!]
\centering
\includegraphics[width=0.9\linewidth]{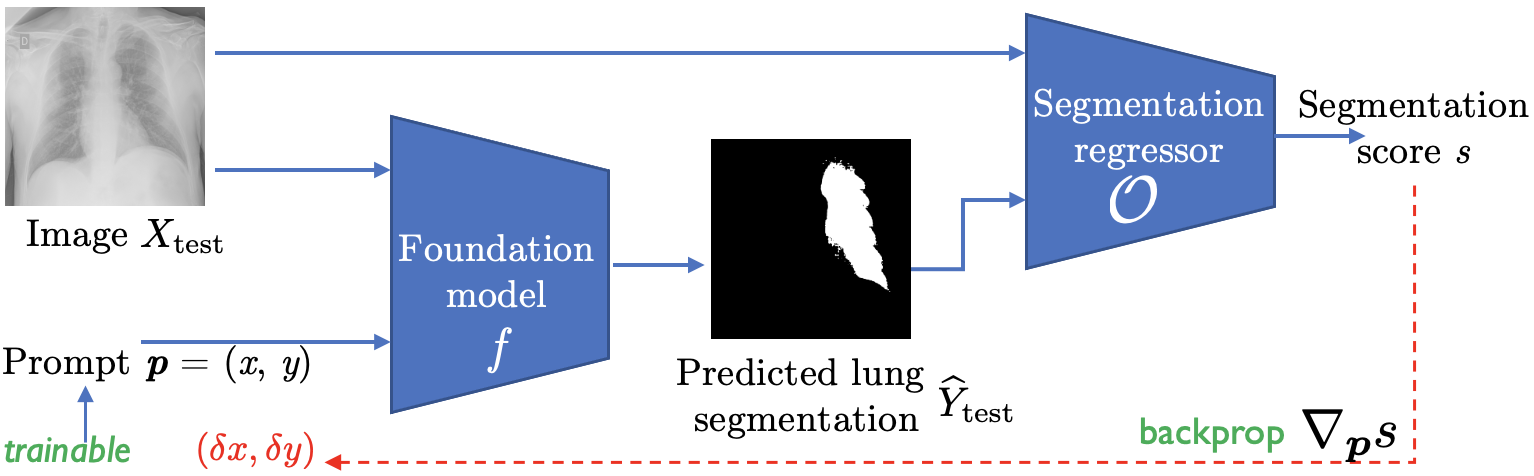}
\caption{Schematic of the SAMPOT. The spatial location of the user-provided prompt is updated based on the gradients received from the segmentation score.}
\label{fig:SAMPOT_scheme}
\end{figure}

\Cref{fig:SAMPOT_scheme} shows the schematic of the proposed SAMPOT approach for prompt learning. Starting from an initial location and an input image, the prompt is iteratively evolved by updating its spatial location using the gradient computed from the segmentation score.

\section{Experiments and Results}
\label{sec:expt}
\subsection{Dataset description}
In this study, we tapped into a database of X-ray images available within our institution, sourced through data partnerships from US, Africa, and European populations. The datasets were acquired after receiving approval from the relevant Institutional Review Boards. The lung boundaries on the X-ray images were delineated by a team of experienced radiologists. X-ray images from $122$ subjects were split into train and test subjects in our experimental setup. This split was used for training and evaluation of the segmentation regressor only. Note that the SAM model is pretrained and is fixed throughout the study. We have evaluated the effectiveness of the prompt optimization technique on the test split of the dataset, thereby ensuring that the results are not biased by the regressor which has been optimized on the train split. The train cohort is further divided into training and validation sets with images from $41$ and $28$ subjects each. The test set has images from $53$ subjects.

\subsection{Segmentation Regressor}
\label{sec:seg_reg}
\begin{figure}[!t]
\subfigure[Sample Mask]{\includegraphics[height=0.19\textwidth]{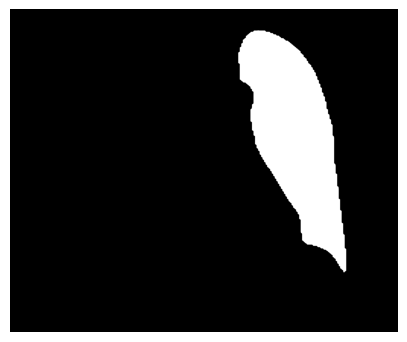}}
\subfigure[Distance map]{\includegraphics[height=0.19\textwidth]{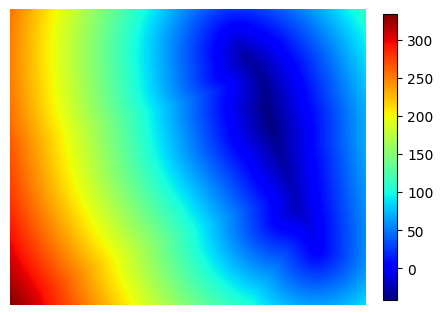}}
\subfigure[Over-seg.]{\includegraphics[height=0.19\textwidth]{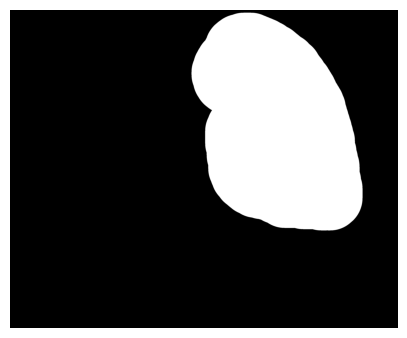}}
\subfigure[Under-seg.]{\includegraphics[height=0.19\textwidth]{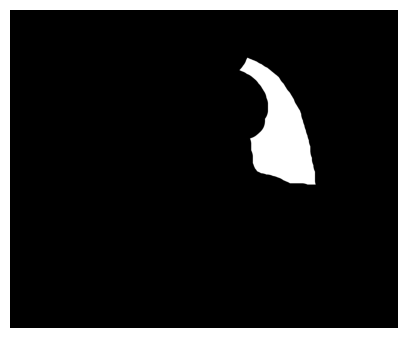}}
\caption{Figure shows (a) sample mask from the dataset, (b) computed distance map,  synthetically generated (c) over-segmented mask and (d) sample under-segmented. The dice coefficient for the over-segmented mask is $0.57$ and that for under-segmented mask is $0.61$.}
\label{fig:reg_data}
\end{figure}

\noindent{\textbf{Data preparation:}} We created several synthetic masks for every lung annotation in the dataset and computed the Dice coefficient for these masks as the ground truth segmentation score. We used the level-sets of ground truth annotation to generate under- and over-segmented instances of the lung field as presented in~\cref{fig:reg_data}. 

Additionally, we also included the lung mask predicted by the SAM when given a single positive prompt and the corresponding Dice coefficient. In every image, the lung field was divided into three horizontal bands and the centroid of these regions were chosen as a prompt. We also chose random points outside the three bands, with an offset of $5$ pixels as prompts for SAM. Therefore, we obtained predictions corresponding to $6$ separate prompts for each image. Thus we had a total of $901$ images in the train set, $600$ in the val set and $1205$ in the test set for learning the regressor.

\noindent{\textbf{Training parameters and network architecture:}} 
The regressor network consisted of five 2D convolution layers interleaved with Batch normalization and leaky ReLU activation, and sigmoid activation for the final layer. The network was trained for $200$ epochs with a batch size of $32$ using Adam optimizer and mean squared error (MSE) loss. A constant learning rate of $0.001$ was used. We set the stopping criterion as minimal loss on the validation set. 
\subsection{Prompt optimization}
Under the mild assumption that a human end-user would choose a prompt located centrally within the region of interest, we chose the centroid of the lung mask as the initial prompt to mimic the human user. Subsequently, the optimization of the prompt location was carried out using Adam optimizer. The step size for the prompt update was heuristically chosen as $10$ and the weight decay was set to zero. To ensure that the input to the regressor (SAM prediction) is closer to a binary mask, we employed sigmoid activation a with steeper slope. Furthermore, we chose the optimal prompt as the one that maximized the output of the regressor. We have used ViT-B SAM in our experiments.

\subsection{Results}
\subsubsection{Evaluation of Segmentation Regressor:}
\begin{figure}
\centering
\subfigure[]{\includegraphics[width=0.32\linewidth]{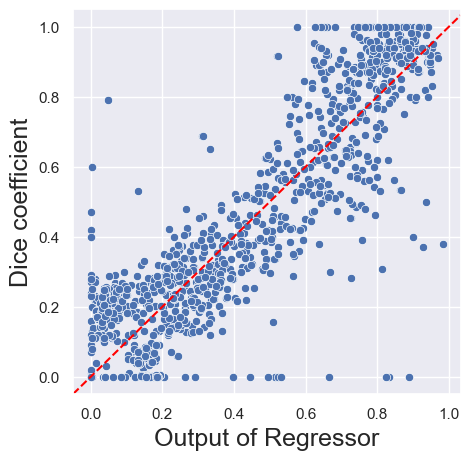}\label{fig:reg_corr}}
\subfigure[]{\includegraphics[width=0.32\linewidth]{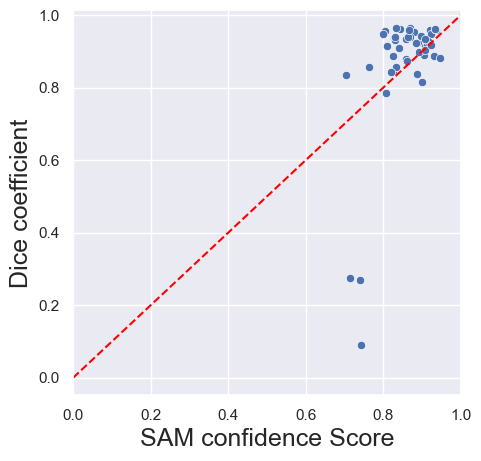}\label{fig:sam_corr}}
\subfigure[]{\includegraphics[width=0.32\linewidth]{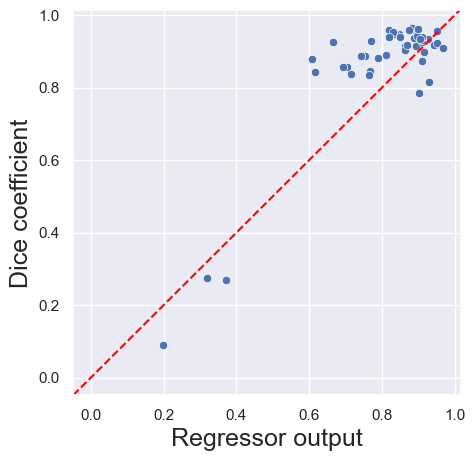}\label{fig:reg_centroid_corr}}

\caption{Comparison of (a) Dice against regressor output for unseen synthetically generated masks (1205 samples); on the test set (53 samples) (b) Dice against SAM confidence score and (c) Dice against regressor output when prompts are placed at the centroid of the lung mask. The correlation coefficient for the regressor on unseen synthetically generated masks is $0.88$. On test samples, the correlation coefficient for the regressor is $0.90$ in comparison with $0.67$ for SAM.}
\label{fig:dice_reg_corr}
\end{figure}

\Cref{fig:reg_corr} is the scatterplot of regressor outputs against true Dice coefficients for all the samples in the test set, including the synthetically generated masks as well as SAM predictions. The high correlation coefficient ($0.88$) shows that the regressor output can serve as a proxy for Dice coefficient of segmented mask. We also present a similar plot for SAM confidence scores for segmentations when prompted at the centroid of the lung mask. We observe that the confidence scores of SAM have a lower correlation coefficient of $0.67$ with Dice compared to our Segmentation Regressor.

\subsubsection{Evaluation of prompt optimization:}
\begin{figure}[t!]
\centering
\includegraphics[width=0.8\linewidth]{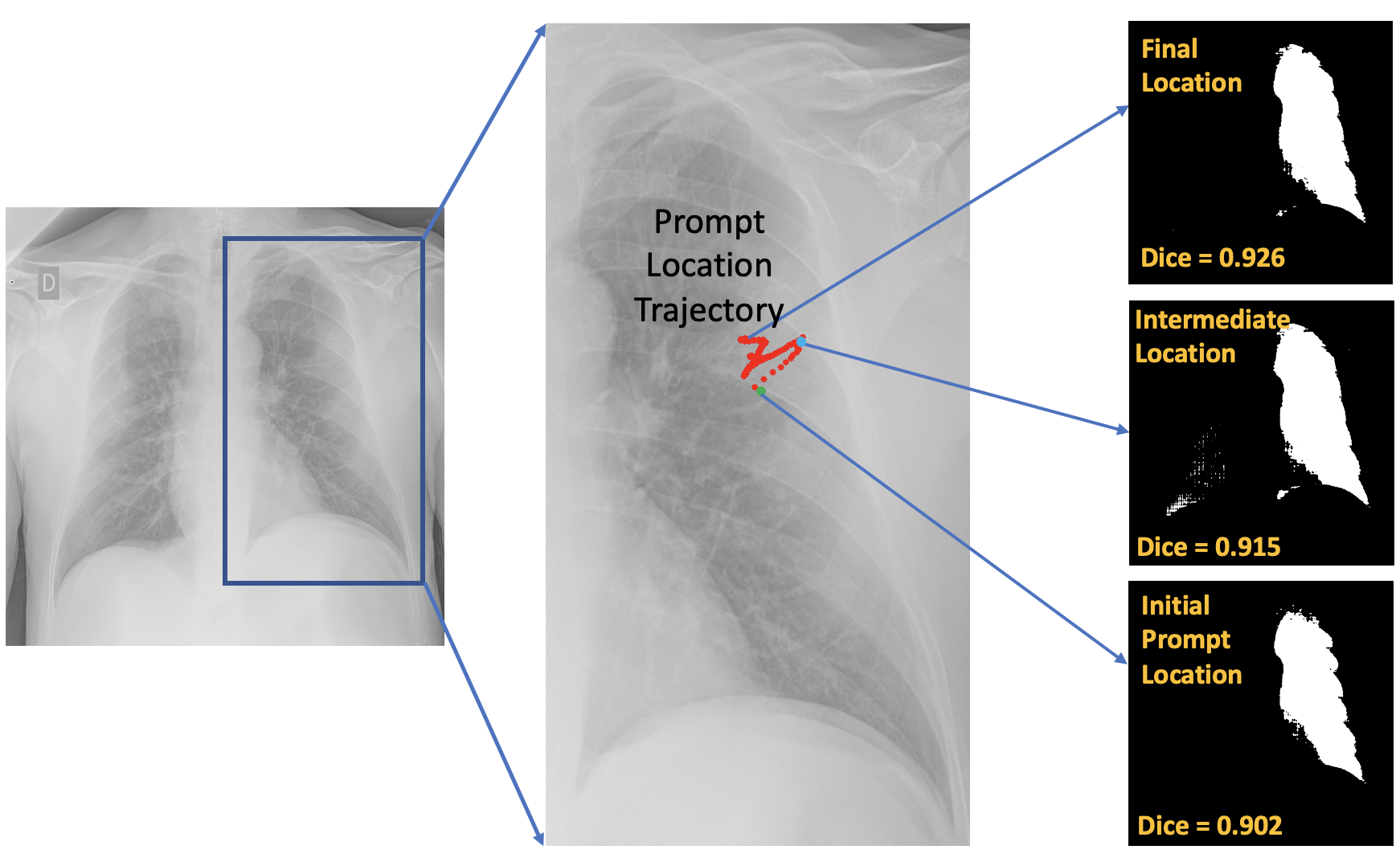}
\caption{[Best viewed in color] Figure illustrates the trajectory of the prompt during the optimization process. The initial prompt is set at the centroid of the ground truth lung field annotation. Snapshots of the predicted masks at select locations on the prompt trajectory along with the computed dice score are also shown.}
\label{fig:prompt_trajectory}
\end{figure}

An illustration of the prompt optimization process for a sample image, starting from the initial location to the optimal location on the image is presented in~\cref{fig:prompt_trajectory}. We see how the quality of the predicted lung field mask, measured using Dice coefficient, improves as the prompt traverses through the optimization trajectory.

\begin{figure}[!th]
\centering
\includegraphics[width=\linewidth]{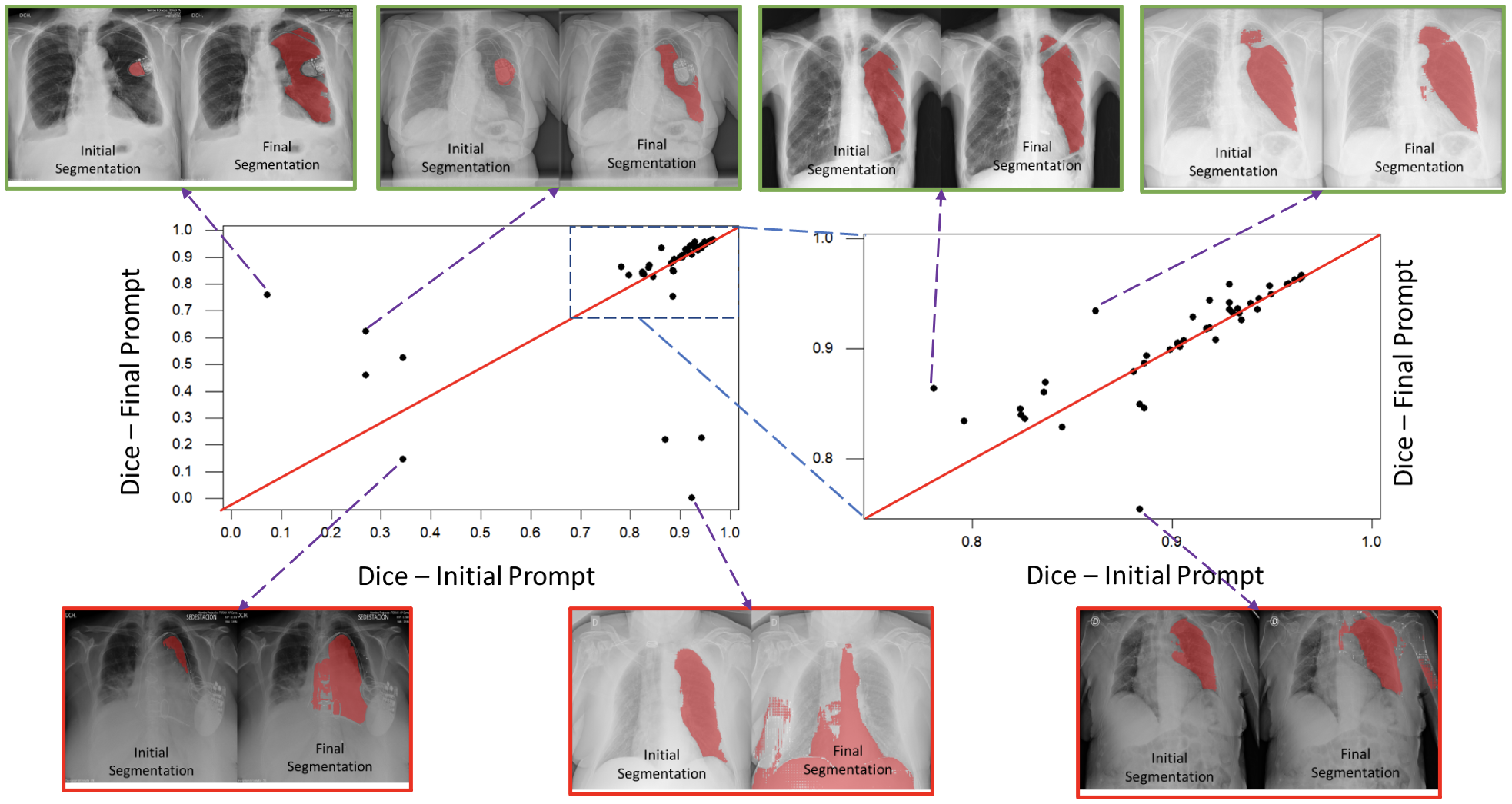}
\caption{[Best viewed in color] Scatter plot of Dice coefficients resulting from initial prompts and the final evolved prompts on the test set. 38 of 53 cases have shown improvement in Dice after evolving. Four of them have significant Dice gains. The scatter plot on the right is the blown-up area on the top-right of the scatter plot on the left. The top row shows images that have significantly gained from prompt evolution. On the bottom are some cases which under-performs upon prompt evolution.}
\label{fig:dice_comparison}
\end{figure}

\Cref{fig:dice_comparison} summarizes the overall performance of the proposed SAMPOT on the test dataset. The scatterplot on the left (initial Dice vs Dice after evolution) shows that $38$ of $53$ images have improved Dice (points above unit slope line) after prompt evolution. Of them, four images have significant improvements. The scatter plot on the right is a blown-up version of a portion of the scatter plot on the left. The images on the top row contain specific examples where the Dice improved after evolution. On the bottom row, the images contain examples of underperforming cases. For the first two under-performing cases displayed, the segmentation masks after evolution are outside the lung region, even though the initial masks were in the right places. Such catastrophic cases can be handled by employing additional safeguard logic.

\section{Discussion}
\label{sec:disc}
The direct application of foundation models like SAM has shown sub-par performance on a number of different medical image segmentation tasks. Given the relatively modest sizes of datasets available for downstream medical imaging tasks, it may be prohibitive to fine-tune a very large model like SAM. The performance of SAM  on the previously unseen problem of lung segmentation on X-ray images is elevated by SAMPOT indicating the possibility of deploying SAM on medical image segmentation problems even with few images.

While this work focused only on prompt evolution, the idea of adapting the input to improve the performance of a foundation model is very generic. One can adapt the input image itself, along with the prompt, to meet the desired objective. A future extension to this work can be adaptation to cases where multiple heterogeneous prompts such as bounding boxes, text inputs etc. are optimized. An extensive evaluation of SAMPOT on a multitude of datasets/use-cases will be beneficial as well. 

\section{Conclusions}
\label{sec:conc}
On medical images, we observed that the spatial location of the prompt for a  general purpose foundation model (SAM) affects the accuracy. Taking a cue from the NLP community, we have presented SAMPOT, a method to optimize the prompt for a foundation model by altering the spatial location to obtain superior results on downstream tasks. We have demonstrated this method on lung segmentation of chest X-rays and obtained improvement on a significant number of cases ($\sim75\%$). We hope that our work offers possibilities of prompt-learning for extracting maximal value from general purpose foundation models trained on natural images on domain-specific downstream tasks in medical image analysis.
%
%
%
\bibliographystyle{splncs04}
\bibliography{mybibliography}
\end{document}